\documentclass{ecai} 

\usepackage[utf8]{inputenc}
\usepackage[T1]{fontenc}
\usepackage[export]{adjustbox}
\usepackage{booktabs}
\usepackage{enumitem}
\usepackage{graphicx}
\usepackage{latexsym}
\usepackage{makecell}
\usepackage{microtype}
\usepackage{multicol}
\usepackage{multirow}
\usepackage{nicefrac}
\usepackage{subcaption}
\usepackage{url}
\usepackage{xcolor} 

\usepackage[mathscr]{euscript}
\usepackage{algorithm}
\usepackage{algpseudocode}
\usepackage{amsfonts}
\usepackage{amsmath}
\usepackage{amssymb}
\usepackage{amsthm}
\usepackage{bm}
\usepackage{mathtools}
\usepackage{listings}
\usepackage{cleveref}

\newtheorem{definition}{Definition}

\newcommand{\BibTeX}{B\kern-.05em{\sc i\kern-.025em b}\kern-.08em\TeX}

\begin{document}


\begin{frontmatter}


\paperid{234} 


\title{Enhancing Node Representations for Real-World Complex Networks with Topological Augmentation}


\author[A]{\fnms{Xiangyu}~\snm{Zhao}\thanks{Equal contribution.}\footnote{Corresponding authors. Email: \{x.zhao22,zehui.li22\}@imperial.ac.uk.}}
\author[A]{\fnms{Zehui}~\snm{Li}\footnotemark[*]\footnotemark[1]}
\author[A]{\fnms{Mingzhu}~\snm{Shen}}
\author[A]{\fnms{Guy-Bart}~\snm{Stan}}
\author[B]{\fnms{Pietro}~\snm{Li\`{o}}}
\author[A]{\fnms{Yiren}~\snm{Zhao}}

\address[A]{Imperial College London, United Kingdom}
\address[B]{University of Cambridge, United Kingdom}


\begin{abstract}
Graph augmentation methods play a crucial role in improving the performance and enhancing generalisation capabilities in Graph Neural Networks (GNNs). Existing graph augmentation methods mainly perturb the graph structures, and are usually limited to pairwise node relations. These methods cannot fully address the complexities of real-world large-scale networks, which often involve higher-order node relations beyond only being pairwise. Meanwhile, real-world graph datasets are predominantly modelled as simple graphs, due to the scarcity of data that can be used to form higher-order edges. Therefore, reconfiguring the higher-order edges as an integration into graph augmentation strategies lights up a promising research path to address the aforementioned issues. In this paper, we present \emph{Topological Augmentation (TopoAug)}, a novel graph augmentation method that builds a combinatorial complex from the original graph by constructing virtual hyperedges directly from the raw data. TopoAug then produces auxiliary node features by extracting information from the combinatorial complex, which are used for enhancing GNN performances on downstream tasks. We design three diverse virtual hyperedge construction strategies to accompany the construction of combinatorial complexes: (1) via graph statistics, (2) from multiple data perspectives, and (3) utilising multi-modality. Furthermore, to facilitate TopoAug evaluation, we provide 23 novel real-world graph datasets across various domains including social media, biology, and e-commerce. Our empirical study shows that TopoAug consistently and significantly outperforms GNN baselines and other graph augmentation methods, across a variety of application contexts, which clearly indicates that it can effectively incorporate higher-order node relations into the graph augmentation for real-world complex networks. 
\end{abstract}

\end{frontmatter}


\section{Introduction}

\begin{figure*}[t]
    \centering
    \includegraphics[width=\textwidth]{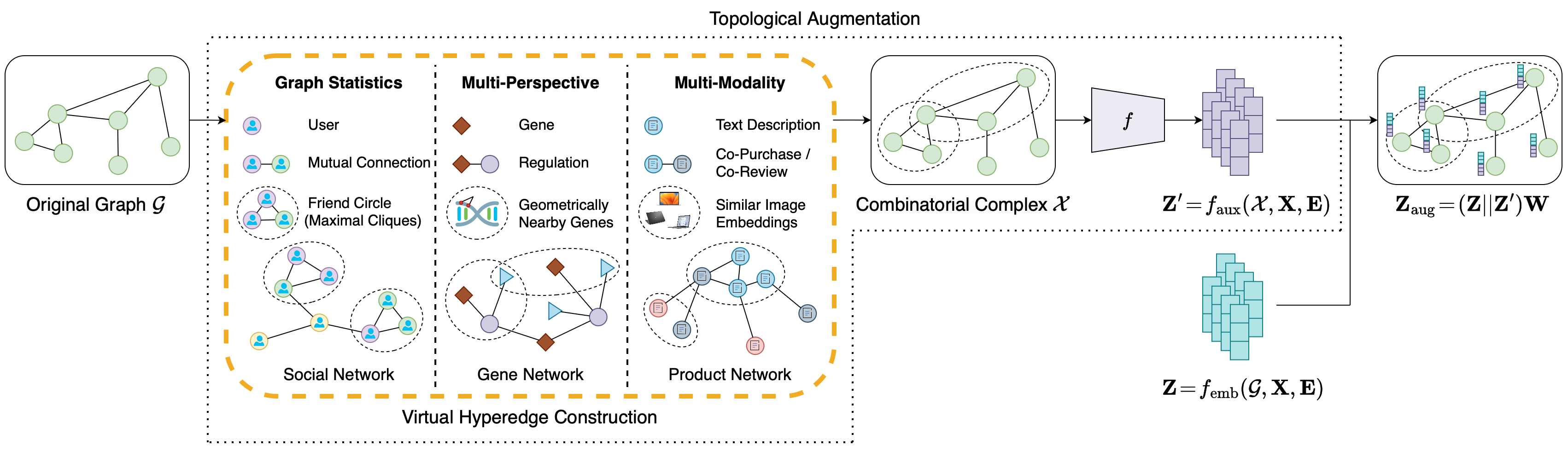}
    \caption{An overview of the TopoAug method showcasing three virtual hyperedge construction strategies: via graph statistics, from multiple data perspectives, and using multiple data modalities. The process initiates with an original graph $\mathcal{G}$. The virtual hyperedge construction methods contribute to the construction of an combinatorial complex $\mathcal{X}$, which is then processed through an auxiliary function $f_\text{aux}$ to obtain a set of auxiliary node features $\mathbf{Z}'$. The auxiliary features $\mathbf{Z}'$ are then combined with the unaugmented node embeddings $\mathbf{Z}$ (obtained from the original graph via the backbone embedding function $f_\text{emb}$), to produce the final augmented node embeddings $\mathbf{Z}_\text{aug}$. The final augmented node embeddings $\mathbf{Z}_\text{aug}$ are then used in downstream tasks to enhance prediction accuracy.}
    \vspace{\baselineskip}
    \label{fig:topoaug}
\end{figure*}

Graph Neural Networks (GNNs)~\cite{brody2021attentive,hamilton2017inductive,kipf2016semi,velivckovic2017graph,zeng2019graphsaint} are powerful tools for learning the representation of relationships between objects, ranging from social networks~\cite{Monti2019FakeNews}, biology~\cite{Rozemberczki2021MUSAE} to e-commerce~\cite{he2016ups,mcauley2015image,ni2019justifying}. Representation learning tasks, such as node prediction, constitute a major category of tasks for GNNs.
However, limitations in the generalisability of GNNs have obstructed their broader application in real-world settings~\cite{rusch2023survey}. 

Several graph augmentation techniques have been introduced to enhance the performance and foster better generalisation for GNNs, specifically in the context of node prediction tasks involving large graphs. Current graph augmentation methods include graph structures perturbation \cite{ding2022data,rong2019dropedge}, feature perturbation \cite{feng2019graph}, and label-oriented augmentations \cite{han2022gmixup,wang2021mixup}. Simple-GNNs -- GNNs that operate on simple graphs with only pairwise relations -- normally use perturbation-based augmentation techniques. However, these augmentation methods do not adequately grasp the complexities of real-world, large-scale networks that frequently display intricate node relationships beyond having only pairwise connections.
While higher-order GNNs, such as Hypergraph Neural Networks (hyper-GNNs) \cite{dong2020hnhn,Wang2023EDHNN,yadati2019hypergcn}, have been developed to model higher-order node relations, they encounter a significant challenge due to the scarcity of explicitly recorded data for forming high-order edges in many real-world datasets. This scarcity hampers the direct application of hyper-GNNs, as they rely on comprehensive data that capture complex node relations. 

Although the scarcity of higher-order edges presents a challenge, integrating advanced modeling methods into augmentation strategies offers a promising avenue to address this issue.
Traditionally, graph augmentation has focused on first-order connections, primarily modifying or adding direct links between nodes~\cite{ding2022data,rong2019dropedge}. However, this approach overlooks the rich, multi-level interactions captured in higher-order edges, which encompass indirect connections, such as patterns of group interactions within the networks. By incorporating these higher-order relations, graph augmentation can achieve a more nuanced understanding of network dynamics, revealing hidden structures and patterns that are not apparent in direct links alone. This approach is particularly beneficial in real-world complex networks, where the interplay of various types of connections can provide deeper insights into the underlying system, across a wide range of domains including social media, biological ecosystems and e-commerce networks. Therefore, the integration of higher-order edge information into graph augmentation strategies opens up new possibilities for more sophisticated and accurate network analysis.

In our study, we introduce a suite of graph augmentation techniques that extend beyond conventional approaches. Our methodology encompasses higher-order node relations, and we present a novel graph augmentation strategy termed \emph{Topological Augmentation (TopoAug)}. As illustrated in \Cref{fig:topoaug}, TopoAug employs a two-step augmentation mechanism. First, TopoAug constructs hyperedges from the original graph data, thereby forming a \emph{combinatorial complex}~\cite{hajij2022topological}. Second, TopoAug further processes the combinatorial complex through an auxiliary function ($f_\text{aux}$ in \Cref{fig:topoaug}) to produce auxiliary node embeddings. These auxiliary embeddings can then be used by GNNs to improve their robustness and accuracy on downstream tasks. We present three distinct ways for constructing these hyperedges: (1) via graph statistics, (2) from multiple data perspectives, and (3) using multiple data modalities. Our augmentation schemes showcase notable flexibility across a variety of datasets, catering to diverse applications such as social, biological, e-commerce, and knowledge networks. Our main contributions are as follows: 

\begin{itemize}
    \item We introduce TopoAug, a novel data augmentation method for GNNs on node prediction tasks. TopoAug allows the network to integrate high-order edge information. Along with our method, we also release 23 new datasets, encompassing diverse domains including social media, biology as well as e-commerce networks.
    \item We assess various design choices within TopoAug, including different methods for constructing virtual hyperedges and varying auxiliary functions. Our examination of diverse TopoAug variants across multiple GNNs and prediction tasks underscores TopoAug's versatility for various application contexts. 
    \item Through a comprehensive empirical evaluation on various real-world datasets and GNN architectures, we demonstrate that Topo-Aug offers consistent and significant performance improvements over GNN baselines and existing graph augmentation methods. 
\end{itemize}

\section{Related Work} \label{section:related}

\paragraph{Topological Deep Learning} Topological deep learning extends machine learning models from graph data to data on topological domains, such as hypergraphs and other higher-order graphs, to facilitate the exponential growth in both the amount and the complexity of data for computational analysis. Formally, a (simple) graph $\mathcal{G}=(\mathcal{V},\mathcal{E})$ is a collection of nodes $\mathcal{V}$ and edges $\mathcal{E}\subseteq\mathcal{V}\times\mathcal{V}$ between pairs of nodes. This simple graph abstraction assumes that each edge only connects two nodes. However, as discussed in the previous sections, many real-world networks have more complex node relations than just pairwise relations. A hypergraph is also defined as $\mathcal{G}=(\mathcal{V},\mathcal{E})$, but $\mathcal{E}\subseteq\mathscr{P}(\mathcal{V})\setminus\varnothing$ is now the set of hyperedges, each of which can connect two or more nodes. However, as real-world datasets are predominantly recorded as simple graphs, treating both simple edges and ``true'' hyperedges (edges that connects strictly more than two nodes) as hyperedges altogether also limits their flexibility, due to the scarcity of data that can form true hyperedges. Recently, there emerges a novel type of topological domain -- \emph{combinatorial complexes}, that enrich the hypergraphs by separating simple edges and true hyperedges through the inclusion of hierarchical ranks~\cite{hajij2022topological}:

\begin{definition}[Combinatorial complex]
A \textbf{combinatorial complex} is a triple $(\mathcal{V},\mathcal{X},\mathrm{rk})$ consisting of a node set $\mathcal{V}$, a node relation set $\mathcal{X}\subseteq\mathscr{P}(\mathcal{V})\setminus\varnothing$, and a rank function $\mathrm{rk}:\mathcal{X}\to\mathbb{N}$, such that 
\begin{enumerate}
    \item $\forall v\in\mathcal{V}.\:\{v\}\in\mathcal{X}$; and
    \item the rank function $\mathrm{rk}$ is order-preserving, which means that\\$\forall x,y\in\mathcal{X}.\:x\subseteq y\implies\mathrm{rk}(x)\leq\mathrm{rk}(y)$.
\end{enumerate} 
\end{definition}

For brevity, $\mathcal{X}$ is used as an abbreviated notation for a combinatorial complex $(\mathcal{V},\mathcal{X},\mathrm{rk})$. Typically, the rank of any singleton node relation $\{v\}$ in $\mathcal{X}$ is set to zero, to make it naturally aligned with simple graphs and hypergraphs. The rank function effectively induces a hierarchical structure on $\mathcal{X}$, which can be used to separate simple edges and true hyperedges, by assigning them with different ranks.

\paragraph{Graph Neural Networks for Simple Graphs} Graph Neural Networks (GNNs) on simple graphs encode the nodes through neural networks, and learn the representations of the nodes through message-passing within the graph structure. GCN incorporates the convolution operation in computer vision into GNNs~\cite{kipf2016semi}. 
GAT~\cite{velivckovic2017graph} and GATv2~\cite{brody2021attentive} are another family of GNN variants that focus on improving the expressive power of GNNs by using the attention mechanism. GraphSAGE~\cite{hamilton2017inductive} is a general inductive framework that leverages node information to efficiently generate node embeddings for previously unseen data. GraphSAINT~\cite{zeng2019graphsaint} underscores the significance of graph sampling-based inductive learning -- it utilises diverse graph sampling techniques and illustrates how learning on smaller, sampled graphs can enhance training efficiency, particularly with large graphs.
While these models demonstrate the success on simple graph datasets, they continue to face challenges in their generalisation to unseen data and are susceptible to small variations in graph structures~\cite{cai2020note}. 

\paragraph{Data Augmentation Methods for GNNs} Enhancing the generalisability of GNNs through graph augmentation methods remains a key area of research interest in recent years.
Existing graph augmentation methods introduce perturbations to the graph structure, enabling GNNs trained on these altered graphs to learn and capture invariance.
This includes adding perturbations to the graphs' adjacency matrices through DropEdge~\cite{rong2019dropedge}, randomly removing nodes through DropNode~\cite{ding2022data,You2020GraphCL}, and perturbing node edge features through feature masking. Another line of graph augmentations relies on the generation of synthetic data, which can be achieved through the interpolation of existing data via Mixup~\cite{han2022gmixup,wang2021mixup,Zhang2018Mixup}, or using generative models to enrich the training data~\cite{liu2023data}. However, these existing augmentation strategies do not inherently increase the expressiveness of GNNs. Standard GNNs typically aggregate information from their neighbourhoods, a process that shares limitations with the Weisfeiler-Lehman (1-WL) graph isomorphism test~\cite{hamilton2020graph,Weisfeiler1968WL,Xu2019GIN}. For instance, as depicted in Figure \ref{fig:wl-test}(a), simple-GNNs fail to distinguish between two structurally different graphs with the same node degrees due to this limitation. This research aims to develop augmentation-based methods that address and overcome these constraints, thereby enhancing the capabilities of GNNs beyond their current limits. 

\paragraph{Representation Learning on Hypergraphs} Hypergraphs are designed to capture more complex node relations, where an edge can connect two or more nodes. In general, GNNs for hypergraphs optimise the node representation through a two-step process. Initially, the node embeddings within each hyperedge are aggregated to form a hidden embedding of each hyperedge. Subsequently, the hidden embeddings of hyperedges with common nodes are aggregated to update the representations of their common nodes. Both HGNN~\cite{feng2019hypergraph} and HyperConv~\cite{Bai2021HyperConvAtten} precisely follow this process. The expressiveness of hypergraph GNNs could be enhanced by modifying this procedure. For instance, HyperGCN~\cite{yadati2019hypergcn} refines the node aggregation within hyperedges using mediators~\cite{chan2020generalizing}. HyperAtten \cite{Bai2021HyperConvAtten} uses attention to measure the degree to which a node belongs to a hyperedge. HNHN~\cite{dong2020hnhn} applies nonlinear functions to both node and edge aggregation processes. ED-HNN~\cite{Wang2023EDHNN} approximates continuous equivariant hypergraph diffusion operators on hypergraphs by feeding node representations into the message from hyperedges to nodes.
GNNs designed for hypergraphs are effective in modelling complex networks. However, real-world datasets are predominantly recorded as simple graphs, which has limited the application of hyper-GNNs due to the scarcity of data that can form hyperedges. 
In this study, we propose TopoAug that derives hyperedges directly from the raw data, thereby extending the applicability of hyper-GNNs to cases where only conventional simple graph data are available. By utilising higher-order GNNs on virtually established hyperedges to produce auxiliary features, TopoAug effectively broadens the potential of hyperedges to scenarios previously confined to simple graph data.

\section{Method}

\subsection{Combinatorial Complex Construction} \label{section:cc}

Inspired by the combinatorial complex modelling, TopoAug attempts to capture complex relations in real-world large networks by constructing a combinatorial complex from the original network. Specifically, TopoAug augments a simple graph by constructing a set of hyperedges $\mathcal{E}_h\subseteq\mathscr{P}(\mathcal{V})\setminus\varnothing$ from the original graph, via a hyperedge extraction function $h:\mathfrak{G}\times\mathbb{R}^{|\mathcal{V}| \times d_v}\times\mathbb{R}^{|\mathcal{E}| \times d_e}\to\mathscr{P}(\mathscr{P}(\mathcal{V}))$, where 
$\mathfrak{G}$ denotes the input graph space, 
$\mathbb{R}^{|\mathcal{V}| \times d_v}$ denotes the $d_v$-dimensional node feature space, 
$\mathbb{R}^{|\mathcal{E}| \times d_e}$ denotes the $d_e$-dimensional edge feature space, and
$\mathscr{P}(\mathscr{P}(\mathcal{V}))$ represents the set of all possible collections of hyperedges, which defines the output hyperedge space: 
\begin{equation}
    \mathcal{E}_h=h(\mathcal{G},\mathbf{X},\mathbf{E})
\end{equation}
where $\mathbf{X} \in \mathbb{R}^{|\mathcal{V}| \times d_v}$ denotes the node feature matrix of the graph, with each row $\mathbf{x}_{v} \in \mathbb{R}^{d_v}$ being the $d_v$-dimensional features of node~$v$, and 
$\mathbf{E} \in \mathbb{R}^{|\mathcal{E}| \times d_e}$ denotes the edge feature matrix of the graph, with each row $\mathbf{e}_{e} \in \mathbb{R}^{d_e}$ being the $d_e$-dimensional features of edge $e$. Each hyperedge $e_h \in \mathcal{E}_h$ is a subset of $\mathcal{V}$ containing at least three nodes, thereby capturing complex multi-node relations. TopoAug then constructs the combinatorial complex $(\mathcal{V},\mathcal{X},\mathrm{rk})$ from the original graph $\mathcal{G}=(\mathcal{V},\mathcal{E})$ and extracted hyperedges $\mathcal{E}_h$, abbreviated as $\mathcal{X}$, according to the following rules:
\begin{align}
    &\mathcal{V}\text{ remains unchanged from the original graph}\\
    &\mathcal{X}=\{\{v\}\:|\:v\in\mathcal{V}\}\cup\{\{u,v\}\:|\:(u,v)\in\mathcal{E}\}\cup\mathcal{E}_h\\
    &\forall x\in\mathcal{X}.\:\mathrm{rk}(x)=\begin{cases}
		0 & \text{for }x=\{v\}\text{ where }v\in\mathcal{V}\\
        1 & \text{for }x=\{u,v\}\text{ where }(u,v)\in\mathcal{E}\\
        2 & \text{otherwise (i.e., for }x\in\mathcal{E}_h\text{)}
	\end{cases}
\end{align}
    
Depending on the nature of the original graph, the virtual hyperedges can be constructed in one of the following three ways: 

\begin{figure}[t]
    \centering
    \begin{subfigure}[t]{0.9\columnwidth}
        \includegraphics[width=\columnwidth]{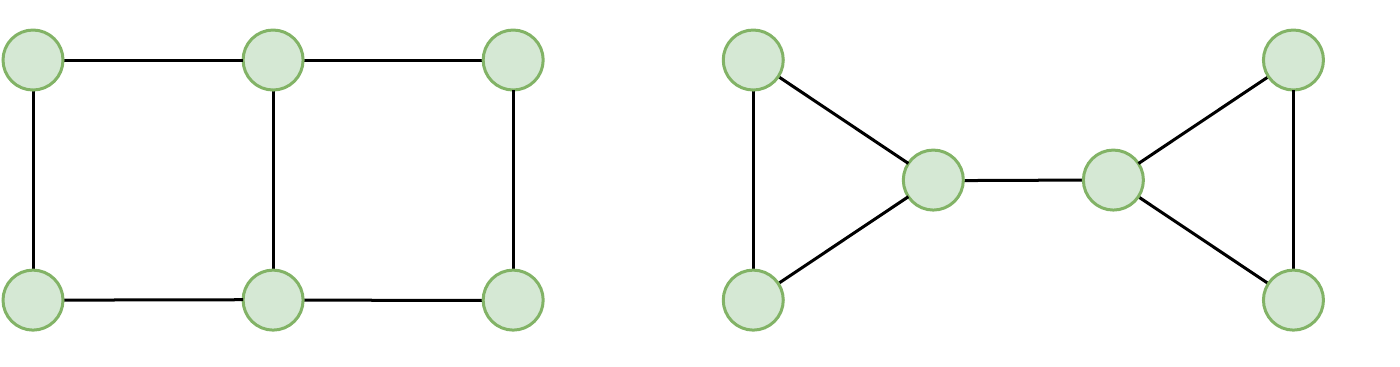}
        \caption{Without TopoAug}
        \vspace{\baselineskip}
    \end{subfigure}
    \begin{subfigure}[t]{0.9\columnwidth}
        \includegraphics[width=\columnwidth]{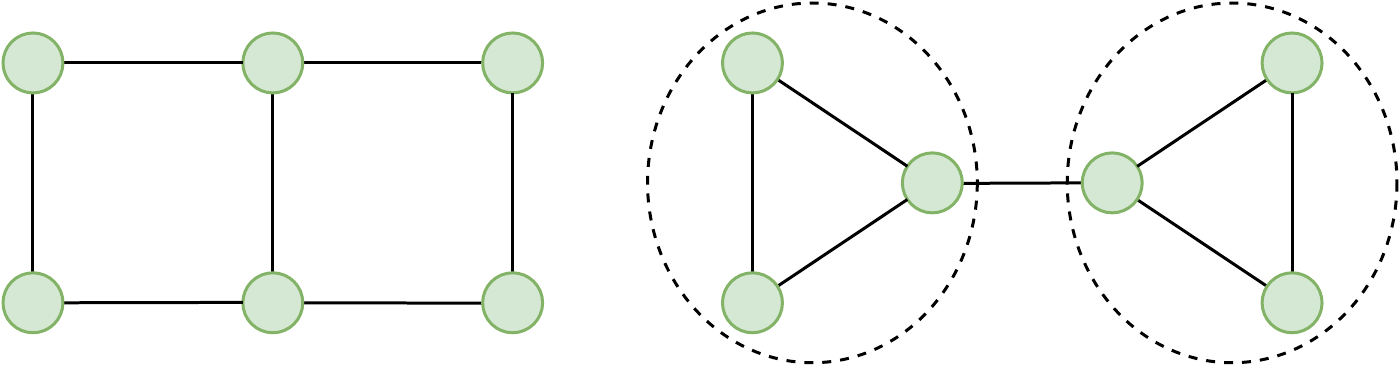}
        \caption{With TopoAug}
        \vspace{\baselineskip}
    \end{subfigure}
    \caption{Graph statistics-based TopoAug can help GNNs surpass the limitations posed by the 1-WL test. (a) Without TopoAug, GNNs are unable to distinguish the two example non-isomorphic graphs that have the same node degrees; (b) Since TopoAug identifies that the right graph contains two cliques, while the left graph contains none, GNNs can now successfully distinguish those graphs.}
    \label{fig:wl-test}
    \vspace{1.5\baselineskip}
\end{figure}

\paragraph{From the Graph Statistics} For graphs that can reveal valuable information from their local clusters and cliques, such as the social networks, TopoAug constructs hyperedges by computing from graph statistics, such as maximal cliques. TopoAug then groups those nodes within the same maximal cliques. This augmentation acts as a complement to the simple edges in the original graph by calculating the local node-wise clustering relationships.
TopoAug then enables GNNs seeking to utilise local clustering information to rapidly access that information without needing to explicitly precompute it.
In addition, by separating out the local clusters and cliques from the simple edge adjacency of the original graph, TopoAug can prevent GNNs from mixing up those information, thereby potentially enhancing its training performance. Another crucial advantage for this type of augmentation is that it enables GNNs to overcome the expressiveness limitations posed by the 1-WL test \cite{hamilton2020graph,Weisfeiler1968WL,Xu2019GIN}: it is able to help GNNs distinguish non-isomorphic graphs with identical node degrees, by specifying their distinct maximal cliques, as illustrated in Figure~\ref{fig:wl-test}.

\paragraph{From a different Data Perspective} Many graphs carry information from different data perspectives. Stacking various types of information from different data perspectives, whether as the graph's node features or edges, can potentially make the learning challenging, since the underlying GNN would then have to learn from these manually mixed information.
Real-world examples of such graphs include the biological networks, where both entity-entity interactions and geometrical information are critical to the prediction tasks. For example, gene regulatory interactions might be represented as an adjacency matrix, while the spatial positioning of genes can be used to establish higher-order relationships. The data inherently offer multiple ways for connecting entities within a graph.
In this scenario, TopoAug constructs hyperedges by extracting information, preferably featuring multi-node relations, from a different data perspective than the node embeddings and edges. These hyperedges can then be processed separately from the node embeddings and edge adjacency, which are derived from a different data perspective, making them ultimately serve as auxiliary node embeddings for the prediction tasks. This type of virtual hyperedge augmentation serves not only as a complement to the simple edges, but also as an enhancement to the overall information obtained -- it takes additional information from a different data source that contain high-order node relations.

\paragraph{From a different Data Modality} For graphs that incorporate information from various data modalities, such as product networks containing both textual and image data, effectively integrating the information from different data modalities can be challenging for a GNN. Similar to the previous scenario, TopoAug addresses this by constructing hyperedges using information from one of the data modalities, preferably exhibiting multi-node relations, which are distinct from the other data modalities used by node embeddings and edges. With this approach, information from different data modalities can be processed independently, and the data from the additional modality essentially serve as an augmentation. Note that this differs from simply incorporating an additional data source into the node embedding, as the alternative data modality may only convey `grouping' information, which is more suitably represented as higher-order graph relations, such as hyperedges.
As the hyperedges are constructed from a different modality, this type of hyperedge augmentation provides both a complement to the simple edges of the original graph, and an enhancement to the overall information.

In topological deep learning, while the simple edges of the graph can be represented as an adjacency matrix $\mathbf{A} \in \{0,1\}^{|\mathcal{V}| \times |\mathcal{V}|}$, where $A_{uv}=1$ if $(u,v) \in \mathcal{E}$ and 0 otherwise, the hyperedges, as well as the final augmented combinatorial complex, can be represented as incidence matrices $\mathbf{H} \in \{0,1\}^{|\mathcal{V}| \times |\mathcal{E}_h|}$ and $\mathbf{H} \in \{0,1\}^{|\mathcal{V}| \times |\mathcal{X}|}$ respectively, with each entry $H_{ve_h}\:(H_{vx})=1$ if $v\in e_h\:(x)$ and 0 otherwise. The combinatorial complex $\mathcal{X}$ is then further processed to produce the final auxiliary features for the original graph's nodes. 

\subsection{Utilising the Hyperedge Information} \label{section:pipeline}

In order to thoroughly and appropriately utilise the additional information from TopoAug's combinatorial complex, we introduce a node feature augmentation pipeline consisting of:

\begin{enumerate}
    \item A combinatorial complex construction function $f_\text{CC}:\mathfrak{G}\to\mathfrak{X}$ mapping from the graph space $\mathfrak{G}$ to the combinatorial complex space $\mathfrak{X}$. $f_\text{CC}$ also encapsulates a hyperedge construction function $h$ (defined in Section~\ref{section:cc}); followed by
    \item A node embedding function $f_\text{emb}:\mathfrak{G}\times\mathbb{R}^{|\mathcal{V}| \times d_v}\times\mathbb{R}^{|\mathcal{E}| \times d_e}\to\mathbb{R}^{|\mathcal{V}| \times d_z}$ that operates on the original graph, together with its unaugmented node and edge features, to compute the $d_z$-dimensional original node embeddings; and
    \item An auxiliary function $f_\text{aux}:\mathfrak{X}\times\mathbb{R}^{|\mathcal{V}| \times d_v}\times\mathbb{R}^{|\mathcal{E}| \times d_e}\to\mathbb{R}^{|\mathcal{V}| \times d_{z'}}$ that operates on the constructed combinatorial complex, together with the original graph's unaugmented node and edge features to produce the $d_{z'}$-dimensional auxiliary node features.
\end{enumerate}

The pipeline outputs the final, $d_{z_\text{aug}}$-dimensional augmented node embeddings as defined in the following equations:
\begin{equation}
    \text{TopoAug}(\mathcal{G}, \mathbf{X}, \mathbf{E}) = (f_\text{emb}(\mathcal{G}, \mathbf{X}, \mathbf{E})\:\|\:f_\text{aux}(f_\text{CC}(\mathcal{G}), \mathbf{X}, \mathbf{E}))\mathbf{W}
\end{equation}
where 
$\mathbf{W} \in \mathbb{R}^{(d_z+d_{z'}) \times d_{z_\text{aug}}}$ denotes the weight matrix of the output layer, and
$\|$ denotes column-wise concatenation.
The final augmented node embeddings are then used for the downstream prediction tasks.
The complete TopoAug pipeline is illustrated in Figure~\ref{fig:topoaug}. While this flexible TopoAug pipeline allows other mechanisms to integrate the auxiliary features into the original graph than a simple concatenation followed by an output layer, it is worth pointing out that this work primarily focuses on exploring the potential of leveraging GNN performance through incorporating higher-order node relations into the original graphs. Designing fine-grained feature integration mechanisms is not the main focus of this research.

\vspace*{-0.75\baselineskip}
\section{Experiments}

\subsection{Datasets} \label{section:data}

\begin{table*}[ht]
    \centering
    \caption{Aggregated statistics of the datasets used for TopoAug evaluation.} 
    \label{table:dataset_stats}
    \begin{tabular}{llrrrrrrr}
        \toprule
        Name & \makecell[l]{Hyperedge\\Construction\\Mechanism} & \#Datasets & \makecell[r]{Avg.\\\#Nodes} & \makecell[r]{Avg.\\\#Edges} & \makecell[r]{Avg.\\\#Hyperedges} &\makecell[r]{Avg.\\Node\\Degree} & \makecell[r]{Avg.\\Hyperedge\\Degree} & \#Classes \\
        \midrule
        MUSAE-GitHub     & Graph Statistics  & 1 & 37,700 & 578,006 & 223,672 & 30.7 &  4.6 & 4 \\
        MUSAE-Facebook   & Graph Statistics  & 1 & 22,470 & 342,004 & 236,663 & 30.4 &  9.9 & 4 \\
        MUSAE-Twitch     & Graph Statistics  & 6 &  5,686 & 143,038 & 110,142 & 50.6 &  6.0 & 2 \\
        MUSAE-Wiki       & Graph Statistics  & 3 &  6,370 & 266,998 & 118,920 & 88.8 & 14.4 & Regression \\
        \midrule
        GRAND-Tissues    & Multi-Perspective & 6 &  5,931 &   5,926 &  11,472 &  2.0 &  1.3 & 3 \\
        GRAND-Diseases   & Multi-Perspective & 4 &  4,596 &   6,252 &   7,743 &  2.7 &  1.3 & 3 \\
        Cora             & Multi-Perspective & 2 & 2,708 & 5,429 & 1,326 & 4.0 & 3.5 & 7 \\
        Pubmed           & Multi-Perspective & 1 & 19,717 & 44,338 & 7,963 & 4.5 & 4.3 & 3 \\
        \midrule
        Amazon-Computers & Multi-Modality    & 1 & 10,226 &  55,324 &  10,226 & 10.8 &  4.0 & 10 \\
        Amazon-Photos    & Multi-Modality    & 1 &  6,777 &  45,306 &   6,777 & 13.4 &  4.8 & 10 \\
        \bottomrule
    \end{tabular}
\end{table*}

To the best of our knowledge, there is a deficiency in the quantity and variety of widely-accepted graph datasets that support data augmentation beyond simple graphs. This is largely attributable to the scarcity of explicitly labelled higher-order node relationships.
In order to thoroughly evaluate the effectiveness of TopoAug across varied domains, we build \emph{23 novel graph datasets} derived from real-world networks across varied domains, including social media, biology, and e-commerce. Care has been taken to ensure that the datasets do not contain any personally identifiable information. The 23 datasets can be split into three groups according to the intended virtual hyperedge construction process for TopoAug:

\paragraph{MUSAE} We build eight social networks derived from the Facebook pages, GitHub developers and Twitch gamers, plus three English Wikipedia page-page networks on specific topics (chameleons, crocodiles and squirrels) based on MUSAE~\cite{Rozemberczki2021MUSAE}. Nodes represent users or articles, and edges are mutual followers relationships between the users, or mutual links between the articles. These datasets are intended to assess TopoAug's effectiveness in constructing virtual hyperedges from the graph statistics: on these datasets, TopoAug constructs the virtual hyperedges to be mutually connected sub-groups that contain at least three nodes (i.e., maximal cliques with sizes of at least 3). The tasks for the Facebook, GitHub, and Twitch datasets involve multi-class classification to predict the categories of users or pages, while the task for the Wiki dataset is a regression task that predicts the average monthly traffic of a web page.

\paragraph{GRAND} We select and build ten gene regulatory networks in different tissues and diseases from GRAND~\cite{ben2022grand}, a public database for gene regulation. Nodes represent gene regulatory elements~\cite{maston2006transcriptional} with three distinct types: protein-encoding gene, lncRNA gene~\citep{long2017lncrnas}, and other regulatory elements. Edges are regulatory effects between genes. We train a CNN~\citep{eraslan2019deep} and use it to take the gene sequence as input and create a 4,651-dimensional embedding for each node. These datasets are intended to assess TopoAug's effectiveness in constructing virtual hyperedges from a different data perspective: on these datasets, TopoAug constructs the virtual hyperedges by grouping geometrically nearby genomic elements on the chromosomes, i.e., the genomic elements within 200k base pair distance. The task is a multi-class classification of gene regulatory elements. 

\paragraph{Amazon} Following existing works on graph representation learning on e-commerce networks~\cite{shchur2018pitfalls,zeng2019graphsaint}, we faithfully reconstruct a subset of the OGB~\cite{Hu2020OGB} \texttt{ogbn-products} dataset, and build two product co-purchase/co-review networks based on the Amazon Product Reviews dataset~\cite{he2016ups,mcauley2015image,ni2019justifying}. Nodes represent products, and an edge between two products is established if a user buys or writes reviews for both products. Node features are extracted based on the textual description of the products. These datasets are intended to assess TopoAug's effectiveness in constructing virtual hyperedges from a different data modality: on these datasets, TopoAug introduces the image modality into the construction of virtual hyperedges. To be specific, the raw images of the products are fed into a CLIP~\cite{radford2021learning} classifier, and a 512-dimensional feature embedding for each image is returned to assist the clustering. TopoAug then constructs the virtual hyperedges by grouping products whose image embeddings have pairwise distances within a certain threshold. The task is to predict the sub-category of a product in a multi-class classification setup.

\medbreak
In addition to the 23 novel datasets, we also adopt three commonly used citation datasets: Cora-CoCitation, Cora-CoAuthorship, and Pubmed-CoCitation, to match TopoAug's performance with the community standard. We use these three datasets to also assess TopoAug's effectiveness in constructing virtual hyperedges from a different data perspective: the edges of the original graph and the virtual hyperedges constructed by TopoAug are co-citation links and co-authorship groups respectively, or vice versa (i.e., co-authorship links and co-citation groups). Table~\ref{table:dataset_stats} reports the key graph statistics for each dataset group, and more details of the datasets are described in Appendices \ref{appendix:all-graph-stats} and \ref{appendix:dataset-details}. We make our source code and full datasets publicly available at \url{https://github.com/VictorZXY/TopoAug}. 

\vspace*{-0.5\baselineskip}
\subsection{Experimental Setup}

\paragraph{Training Details} We run all the experiments on NVIDIA A100 and V100 GPUs, with up to 40GB memory. Adam~\cite{kingma2014method} is used as the optimiser, and CosineAnnealingLR~\cite{gotmare2018closer} is used as the learning rate scheduler for all training. For each experiment, the nodes of the graph dataset are randomly split into training, validation, and test sets with a split ratio of 6:2:2. All models are trained for 500 epochs. For node classification tasks, the negative log likelihood loss (NLLLoss) is used as the loss function. For node regression tasks, the mean square error (MSELoss) is used as the loss function. Each experiment typically takes less than 5 minutes to train, when ED-HNN is not incorporated in the model. Due to the significantly larger architecture of ED-HNN, experiments involving it can take up to 2 hours. 

\paragraph{Hyperparameter Settings} We perform a hyperparameter search for the learning rate and dropout rate, while keeping the hidden dimension of the layers fixed as 64. To ensure fair comparison, all evaluated GNNs (GCN, GAT, GraphSAGE, HyperConv, and ED-HNN) share the same hyperparameter combinations. After hyperparameter searching, we adopt the following hyperparameter selections: learning rate = 0.001, and dropout rate = 0.5. For the additional hyperparameters of ED-HNN, we closely adhere to the hyperparameter settings specified in the original ED-HNN paper~\cite{Wang2023EDHNN}, and set the number of all inner multi-layer perceptrons (MLPs) within ED-HNN to 2. 

\subsection{Designing TopoAug}

As described in Section~\ref{section:pipeline}, the TopoAug pipeline consists of a combinatorial complex construction function $f_\text{CC}$ that constructs a combinatorial complex from the original graph, followed by a backbone node embedding function $f_\text{emb}$ that operates on the original graph to produce the unaugmented node embeddings, and an auxiliary function $f_\text{emb}$ that operates on the constructed combinatorial complex to produce the auxiliary node features. This opens up the following questions regarding the practical design of TopoAug:

\vspace*{-0.25\baselineskip}
\begin{itemize}
    \item At what point should the auxiliary node features be integrated with the original node information -- directly at the input side, or within the middle-layer embeddings?
    \item What is the appropriate type, and consequently, the optimal choice of the auxiliary function?
\end{itemize}

\vspace*{-0.25\baselineskip}
We investigate these design choices and analyse the optimal usage of TopoAug through a series of ablation studies.

\subsubsection{Appropriate Phase for Applying TopoAug}

The auxiliary features generated by TopoAug provide significant flexibility regarding the stage of integration, including simultaneous injection with the input node features or blending with the original GNN's activations.
To thoroughly identify the appropriate place for inserting TopoAug's auxiliary features, we compare the performance of TopoAug with the following settings: 

\vspace*{-0.25\baselineskip}
\begin{enumerate}
    \item concatenate TopoAug's auxiliary features directly with the input node features, then proceed the concatenated features with a linear layer, before feeding into the embedding GNN (denoted as TopoAug\textsubscript{($f_\text{aux}$, input)}, where $f_\text{aux}$ is the auxiliary function); 
    \item concatenate TopoAug's auxiliary features with the GNN activations, followed by a linear layer (denoted as TopoAug\textsubscript{($f_\text{aux}$, emb.)}).
\end{enumerate}

\vspace*{-0.25\baselineskip}
We then conducted experiments on the MUSAE-GitHub, GRAND-Brain and Amazon-Computers datasets, each featuring a distinct virtual hyperedge construction strategy, to evaluate these two settings. The augmentations are applied to a GCN~\cite{kipf2016semi}. HyperConv~\cite{Bai2021HyperConvAtten} and ED-HNN~\cite{Wang2023EDHNN} are selected as the auxiliary functions for TopoAug, since HyperConv is one of the most popular hyper-GNN baselines, and ED-HNN represents the current state-of-the-art in hyper-GNNs. 

Table~\ref{table:ablation-input-vs-emb} summarises the results for this set of ablation studies. The results show a consistent and significant increase in accuracy for GCN with TopoAug\textsubscript{($f_\text{aux}$, emb.)} compared to the vanilla GCN and also GCN with TopoAug\textsubscript{($f_\text{aux}$, input)}. This result suggests that inserting TopoAug's auxiliary features at the output embedding phase is the more appropriate choice. 
Intuitively, this is because concatenating the auxiliary features with the node features at the input phase can potentially mix up different types of input information, and confuse the GNN. 

\begin{table}[t]
    \caption{Evaluating different phases for applying TopoAug (TopoAug\textsubscript{($f_\text{aux}$, input)}, TopoAug\textsubscript{($f_\text{aux}$, emb.)}): accuracy (\%) of the TopoAug variants using GCN as the backbone embedding function $f_\text{emb}$, whose auxiliary features are integrated at different stages, compared with the vanilla GCN without any augmentation. GCN with TopoAug\textsubscript{($f_\text{aux}$, emb.)} variants show a consistent and much more significant accuracy enhancement compared with vanilla GCN and GCN with TopoAug\textsubscript{($f_\text{aux}$, input)} variants.}
    \label{table:ablation-input-vs-emb}
    \centering
    \begin{tabular}{llll}
        \toprule
        Method & GitHub & Brain & Computers \\
        \midrule
        GCN baseline & 87.2 $\pm$ 0.0 & 62.5 $\pm$ 0.0 & 75.6 $\pm$ 4.1 \\ 
        \midrule
        TopoAug\textsubscript{(HyperConv, input)} & 86.9 $\pm$ 0.3 & 63.6 $\pm$ 1.3 & 96.0 $\pm$ 0.5 \\
        TopoAug\textsubscript{(ED-HNN, input)} & 87.1 $\pm$ 0.5 & 63.6 $\pm$ 1.3 & 95.8 $\pm$ 0.4 \\
        \midrule
        TopoAug\textsubscript{(HyperConv, emb.)} & 87.2 $\pm$ 0.0 & 63.7 $\pm$ 0.2 & 96.8 $\pm$ 0.5 \\
        TopoAug\textsubscript{(ED-HNN, emb.)} & \textbf{87.4 $\pm$ 0.3} & \textbf{66.7 $\pm$ 2.6} & \textbf{98.1 $\pm$ 0.7} \\
        \bottomrule
    \end{tabular}
    \vspace{0.5\baselineskip}
\end{table}

\begin{table}[t]
    \caption{Evaluating different types for the auxiliary function $f_\text{aux}$: accuracy (\%) of TopoAug with various simple-GNNs and hyper-GNNs as auxiliary functions, using GCN as the backbone embedding function $f_\text{emb}$, compared with the vanilla GCN without any augmentation. The TopoAug variants with hyper-GNNs as auxiliary functions clearly outperform TopoAug with simple-GNNs.}
    \label{table:ablation-simple-vs-hyper-gnn}
    \centering
    \begin{tabular}{llll}
        \toprule
        Method & GitHub & Brain & Computers \\
        \midrule
        GCN baseline  & 87.2 $\pm$ 0.0 & 62.5 $\pm$ 0.0 & 75.6 $\pm$ 4.1 \\ 
        \midrule
        TopoAug\textsubscript{(GAT)} & 86.7 $\pm$ 0.1 & 62.5 $\pm$ 0.3 & 91.3 $\pm$ 0.1 \\
        TopoAug\textsubscript{(GraphSAGE)} & 87.0 $\pm$ 0.4 & 65.3 $\pm$ 0.9 & 93.0 $\pm$ 0.0 \\
        \midrule
        TopoAug\textsubscript{(HyperConv)} & 87.2 $\pm$ 0.0 & 63.7 $\pm$ 0.2 & 96.8 $\pm$ 0.5 \\
        TopoAug\textsubscript{(ED-HNN)} & \textbf{87.4 $\pm$ 0.3} & \textbf{66.7 $\pm$ 2.6} & \textbf{98.1 $\pm$ 0.7} \\
        \bottomrule
    \end{tabular}
    \vspace{0.5\baselineskip}
\end{table}

\subsubsection{Choice of the Auxiliary Function Type} \label{section:ablation-simple-vs-hyper}

It is intuitive to use hyper-GNNs as auxiliary functions ($f_\text{aux}$) in TopoAug, so that hyperedge information can be fully captured. However, it is still worth verifying that hyper-GNNs indeed outperform simple-GNNs as auxiliary functions in TopoAug, in order to prove the efficacy of TopoAug. Therefore, we compare the performance of TopoAug with different classes of GNNs as the auxiliary function, denoted as TopoAug\textsubscript{($f_\text{aux}$)}: (1) hyper-GNNs, namely HyperConv and ED-HNN; (2) simple-GNNs, namely GAT~\cite{velivckovic2017graph} and GraphSAGE~\cite{hamilton2017inductive}. These ablation experiments are again conducted on the MUSAE-GitHub, GRAND-Brain and Amazon-Computers datasets, and GCN remains as the GNN for applying TopoAug. \emph{Starting from this stage, all experiments apply TopoAug's auxiliary features at the output embedding phase.}
Table~\ref{table:ablation-simple-vs-hyper-gnn} reports the results for this set of ablation studies, which clearly show that TopoAug with hyper-GNNs as auxiliary functions outperform TopoAug with simple-GNNs, thereby justifying that TopoAug indeed utilises the virtual hyperedges it constructs.

\subsubsection{Identifying the Best Auxiliary Function}

\begin{table}[t]
    \caption{Evaluating the optimal auxiliary function $f_\text{aux}$ for TopoAug: average accuracy rankings of three vanilla GNNs and TopoAug with various auxiliary functions on different GNNs, across 23 node classification datasets. GraphSAGE with TopoAug using HyperConv as the auxiliary function proves to be the best-performing model.}
    \label{table:ablation-hyper-gnn}
    \centering
    \begin{tabular}{lr}
        \toprule
        Method & Average Ranking \\
        \midrule
        GCN       & 6.96 \\ 
        GAT       & 7.91 \\
        GraphSAGE & 7.09 \\
        \midrule
        GCN+TopoAug\textsubscript{(HyperConv)} & 5.26 \\
        GCN+TopoAug\textsubscript{(ED-HNN)} & 3.09 \\
        \midrule
        GAT+TopoAug\textsubscript{(HyperConv)} & 5.22 \\
        GAT+TopoAug\textsubscript{(ED-HNN)} & 3.04 \\
        \midrule
        GraphSAGE+TopoAug\textsubscript{(HyperConv)} & \textbf{2.26} \\
        GraphSAGE+TopoAug\textsubscript{(ED-HNN)} & 2.74 \\
        \bottomrule
    \end{tabular}
    \vspace{0.5\baselineskip}
\end{table}

Now that the efficacy of TopoAug has been affirmed through the aforementioned ablation studies, we conduct extensive experiments to identify the optimal auxiliary function for TopoAug, denoted as $f_\text{emb}$+TopoAug\textsubscript{($f_\text{aux}$)}. We use GCN, GAT and GraphSAGE as the embedding GNNs $f_\text{emb}$ for applying TopoAug, and test the performance of TopoAug with either HyperConv or ED-HNN as the auxiliary function $f_\text{aux}$, on all 23 node classification datasets we build (details of the datasets are described in Section~\ref{section:data}).

For clarity, we present in Table~\ref{table:ablation-hyper-gnn} the average rankings of those six TopoAug applications, together with the three vanilla GNNs without TopoAug. The results clearly show that all TopoAug applications surpass the performance of the vanilla GNNs, which further validate the superior efficacy of TopoAug. Among those TopoAug variants, GraphSAGE with TopoAug\textsubscript{(HyperConv)} demonstrates to be the most effective combination. In addition, for both GCN and GAT, TopoAug\textsubscript{(ED-HNN)} outperforms TopoAug\textsubscript{(HyperConv)}, indicating a positive correlation between the expressiveness of the auxiliary hyper-GNN and the performance gain provided by TopoAug. This is likely due to the enhanced ability of more expressive hyper-GNNs to extract hyperedge information, leading to more informative auxiliary features.

\begin{table*}[t]
    \caption{Accuracy (\%) of TopoAug and corresponding baselines and existing graph augmentation methods on selected node classification datasets. *The results for GCN and GAT on the Cora-CoCitation dataset are directly taken from the GAT paper~\cite{velivckovic2017graph}. **The results for HyperConv and ED-HNN on the Cora-CoCitation dataset are directly taken from the ED-HNN paper~\cite{Wang2023EDHNN}. The results clearly show superior performance of TopoAug compared to the baseline GNNs and existing graph augmentation methods.}
    \label{table:main-results}
    \centering
    \begin{tabular}{llllllll}
        \toprule
        \multirow{2}{*}{Method} & \multicolumn{2}{c}{Graph Statistics} & \multicolumn{3}{c}{Multi-Perspective} & \multicolumn{2}{c}{Multi-Modality} \\
        \cmidrule(lr){2-3} \cmidrule(lr){4-6} \cmidrule(lr){7-8}
        & GitHub & TwitchDE & \makecell[l]{Cora-\\CoCitation} & Brain & \makecell[l]{Lung\\Cancer} & Computers & Photos \\
        \midrule
        RandomGuess & \textit{25.0} & \textit{50.0} & \textit{14.3} & \textit{33.3} & \textit{33.3} & \textit{10.0} & \textit{10.0}\\
        \midrule
        GCN       & 87.2 $\pm$ 0.0 & 65.5 $\pm$ 0.2 & 81.4 $\pm$ 0.5* & 62.5 $\pm$ 0.0 & 59.6 $\pm$ 0.1 & 75.6 $\pm$ 4.1 & 29.5 $\pm$ 1.7 \\ 
        GAT       & 86.4 $\pm$ 0.1 & 64.5 $\pm$ 0.4 & 83.0 $\pm$ 0.7* & 62.5 $\pm$ 0.1 & 59.6 $\pm$ 0.0 & 74.2 $\pm$ 4.3 & 43.4 $\pm$ 7.4 \\
        GraphSAGE & 87.1 $\pm$ 0.2 & 65.7 $\pm$ 0.1 & 83.2 $\pm$ 0.1 & 61.8 $\pm$ 0.2 & 61.5 $\pm$ 1.5 & 75.0 $\pm$ 1.9 & 36.6 $\pm$ 6.1 \\
        HyperConv & 80.8 $\pm$ 0.1 & 65.4 $\pm$ 0.2 & 79.1 $\pm$ 1.0** & 62.5 $\pm$ 0.0 & 59.3 $\pm$ 0.3 & 84.2 $\pm$ 2.0 & 33.7 $\pm$ 5.9 \\
        ED-HNN    & 86.2 $\pm$ 0.1 & 68.1 $\pm$ 0.6 & 80.3 $\pm$ 1.4** & 66.3 $\pm$ 1.3 & 60.2 $\pm$ 1.4 & 97.3 $\pm$ 0.2 & 78.6 $\pm$ 1.2 \\
        \midrule
        GCN+DropNode & 86.2 $\pm$ 0.1 & 67.9 $\pm$ 0.7 & 85.5 $\pm$ 1.0 & 63.3 $\pm$ 0.2 & 58.2 $\pm$ 1.3 & 91.8 $\pm$ 2.1 & 71.1 $\pm$ 0.5 \\
        GCN+DropEdge & 86.6 $\pm$ 0.2 & 67.7 $\pm$ 1.0 & 86.3 $\pm$ 0.4 & 63.2 $\pm$ 0.4 & 60.7 $\pm$ 0.3 & 92.2 $\pm$ 0.7 & 78.6 $\pm$ 0.6 \\
        GCN+Mixup & 85.8 $\pm$ 0.3 & 67.6 $\pm$ 0.8 & 85.6 $\pm$ 1.4 & 65.0 $\pm$ 1.0 & 59.0 $\pm$ 0.6 & 87.7 $\pm$ 2.9 & 78.0 $\pm$ 0.1 \\
        GCN+NodeFeatureMasking & 85.9 $\pm$ 0.0 & 67.8 $\pm$ 0.2 & 85.2 $\pm$ 0.9 & 64.0 $\pm$ 1.2 & 59.2 $\pm$ 0.8 & 91.7 $\pm$ 0.3 & 80.7 $\pm$ 0.6 \\
        \midrule 
        GraphSAGE+DropNode & 86.2 $\pm$ 0.1 & \textbf{68.3 $\pm$ 0.6} & 86.4 $\pm$ 0.0 & 63.3 $\pm$ 0.1 & 64.5 $\pm$ 0.8  & 95.9 $\pm$ 0.2 & 69.8 $\pm$ 0.5 \\
        GraphSAGE+DropEdge & 86.8 $\pm$ 0.1 & 67.8 $\pm$ 1.3 & 87.1 $\pm$ 0.1 & 64.2 $\pm$ 0.2 & 64.7 $\pm$ 0.7 & 95.9 $\pm$ 0.3 & 80.5 $\pm$ 0.6 \\
        GraphSAGE+Mixup & 85.9 $\pm$ 0.3 & 67.1 $\pm$ 0.2 & \textbf{87.2 $\pm$ 1.3} & 64.2 $\pm$ 1.5 & 63.4 $\pm$ 0.1  & 92.2 $\pm$ 0.0 & 71.5 $\pm$ 0.3 \\
        GraphSAGE+NodeFeatureMasking & 86.1 $\pm$ 0.0 & 68.1 $\pm$ 0.7 & 87.1 $\pm$ 0.5 & 64.4 $\pm$ 0.4 & 64.9 $\pm$ 2.3 & 95.7 $\pm$ 0.2 & 79.3 $\pm$ 0.8 \\
        \midrule 
        GCN+TopoAug & \textbf{87.4 $\pm$ 0.3} & 67.9 $\pm$ 0.9 & 86.6 $\pm$ 1.4 & \textbf{66.7 $\pm$ 2.6} & 63.7 $\pm$ 0.9 & 98.1 $\pm$ 0.7 & \textbf{80.9 $\pm$ 0.3} \\ 
        GraphSAGE+TopoAug & 87.3 $\pm$ 0.2 & \textbf{68.3 $\pm$ 0.3} & \textbf{87.2 $\pm$ 1.8} & 66.6 $\pm$ 1.6 & \textbf{66.4 $\pm$ 0.2} & \textbf{98.2 $\pm$ 0.7} & \textbf{80.9 $\pm$ 0.7} \\ 
        \bottomrule
    \end{tabular}
    \vspace{0.5\baselineskip}
\end{table*}

\begin{table}[t]
    \vspace{0.5\baselineskip}
    \caption{MSE ($\downarrow$) of TopoAug and three vanilla GNNs on the node regression datasets (MUSAE-Wiki). The results clearly show superior performacne of TopoAug compared to the vanilla GNN baselines.}
    \label{table:hyperaug-wiki}
    \centering
    \begin{tabular}{llllll}
    \toprule
        Method & Squirrel & Crocodile & Chameleon \\
        \midrule
        GCN & 7.319 $\pm$ 0.000 & 8.761 $\pm$ 0.001 & 6.779 $\pm$ 0.005 \\
        GAT & 7.313 $\pm$ 0.007 & 8.093 $\pm$ 0.054 & 6.249 $\pm$ 0.261 \\
        HyperConv & 7.230 $\pm$ 0.002 & 8.706 $\pm$ 0.000 & 6.712 $\pm$ 0.001 \\
        \midrule
        GCN+TopoAug & 6.557 $\pm$ 0.154 & \textbf{4.851 $\pm$ 0.014} & 5.515 $\pm$ 0.008 \\
        GAT+TopoAug & \textbf{6.049 $\pm$ 0.204} & 4.875 $\pm$ 0.031 & \textbf{4.665 $\pm$ 0.050} \\
        \bottomrule
    \end{tabular}
    \vspace{0.5\baselineskip}
\end{table}

\subsection{Main Results}

We thoroughly evaluate the performance of TopoAug against three simple-GNN baselines: GCN, GAT and GraphSAGE, and two hyper-GNN baselines: HyperConv and ED-HNN, on all 23 node classification datasets we build. 
We include the two hyper-GNNs as baselines since the augmented combinatorial complexes constructed by TopoAug also make use of the hyperedge information. 

Moreover, we also compare the performance of TopoAug with four different existing graph augmentation methods: DropNode~\cite{You2020GraphCL}, DropEdge~\cite{rong2019dropedge}, Mixup~\cite{Zhang2018Mixup} and NodeFeatureMasking~\cite{feng2019graph}. We evaluate those graph augmentation methods on seven selected datasets: MUSAE-GitHub and MUSAE-TwitchDE (featuring virtual hyperedge construction from graph statistics); Cora-CoCitation, GRAND-Brain, and GRAND-LungCancer (featuring virtual hyperedge construction from different data perspectives); as well as Amazon-Computers and Amazon-Photos (featuring virtual hyperedge construction from different data modalities). Each of dataset chosen in this comparison is among the largest graphs in their corresponding dataset subgroups. GCN and GraphSAGE are used as the GNNs for the evaluation with the graph augmentation methods. Details about the experimental settings and the full results are provided in Appendix~\ref{appendix:results}.

Table~\ref{table:main-results} summarises the results of the experiments. The results unequivocally demonstrate that \emph{TopoAug consistently outperforms~the vanilla GNN baselines and other graph augmentation methods, across all three types of virtual hyperedge construction strategies.} This again highlights the superiority of TopoAug, as well as the idea of incorporating higher-order node relations into graph augmentation methods for real-world complex networks. Besides, TopoAug also outperforms both hyper-GNN baselines, thereby indicating that integrating the original graph with the auxiliary features as augmentations is preferable over only learning from the virtual hyperedges and auxiliary features alone. Among the two backbone embedding GNNs on which TopoAug is applied, GraphSAGE with TopoAug performs better than GCN with TopoAug in general, which is likely due to the more expressive nature of GraphSAGE on those tasks, compared with GCN.

It is noted that on the MUSAE-TwitchDE and Cora-CoCitation datasets, although TopoAug is still the best performing method, its performance improvements compared to other graph augmentation methods are not as significant as on other datasets. This can be attributed to the following specific characteristics of the two datasets: 
\begin{itemize}
    \item The MUSAE-TwitchDE and Cora-CoCitation datasets are comparatively small containing a few thousands of nodes, and most appropriate graph augmentation methods can help the GNNs saturate in extracting their information.
    \item The virtual hyperedges for the MUSAE-TwitchDE dataset are constructed by computing from the original graph's statistics, which does not introduce new information to the graph, thus also limiting TopoAug's effectiveness in leveraging GNNs' performance. 
    \item The virtual hyperedges for the Cora-CoCitation dataset are constructed by grouping papers with co-authors, which are not diverse to the co-citation links and may contain redundant information, compared with the GRAND (gene regulations vs. geometrical information) and Amazon (text vs. image modality) datasets.
\end{itemize}
These findings provide valuable guidance for employing TopoAug in real-world applications, in order to fully unleash its potential.

In addition, we also perform supplementary evaluation of TopoAug on the three node regression datasets: MUSAE-Wiki-Chameleon, MUSAE-Wiki-Crocodile and MUSAE-Wiki-Squirrel, against the GCN, GAT and HyperConv baselines, and report their MSEs in Table~\ref{table:hyperaug-wiki}. The results also show consistently and significantly lower MSEs when TopoAug is applied, which further validate the superiority of TopoAug. On those node regression datasets, GAT with TopoAug generally outperforms GCN with TopoAug, since GAT is more expressive than GCN on those tasks. This observation also aligns well with the results on the node classification datasets, which suggests that TopoAug is stable across various types of tasks. 

\section{Conclusion}

This work introduces \emph{Topological Augmentation (TopoAug)}, a novel graph augmentation method designed to improve the performance of GNNs in node prediction tasks across diverse real-world large-scale graph datasets. TopoAug enhances the capability of GNNs by constructing combinatorial complexes from the original graphs using virtual hyperedges, and then generating auxiliary features for each node. This improvement allows even basic GNN models to surpass the limitations of the 1-WL test. Empirical results confirm that TopoAug outperforms traditional graph augmentation methods in terms of node prediction across various data domains. 

For future work, there are several potential methods that could be explored to further utilise the auxiliary information. These include applying a cross-attention mechanism, as detailed in works~\cite{vaswani2017attention,wei2020multi}, to effectively integrate the auxiliary and original features. Another unexplored avenue is the concurrent optimisation of the auxiliary and original features using contrastive loss, a technique discussed in ViLT~\cite{kim2021vilt}. Employing these methods has the potential to optimise the integration and utility of the auxiliary features within GNNs, thereby further increasing their overall performance.



\begin{ack}
This work was performed using the Sulis Tier 2 HPC platform hosted by the Scientific Computing Research Technology Platform at the University of Warwick, the JADE Tier 2 HPC facility, and the Cirrus UK National Tier-2 HPC Service at EPCC. Sulis is funded by EPSRC Grant EP/T022108/1 and the HPC Midlands+ consortium. JADE is funded by EPSRC Grant EP/T022205/1. Cirrus is funded by the University of Edinburgh and EPSRC Grant EP/P020267/1. Xiangyu Zhao acknowledges the funding from the Imperial College London Electrical and Electronic Engineering PhD Scholarship. Zehui Li acknowledges the funding from the UKRI 21EBTA: EB-AI Consortium for Bioengineered Cells \& Systems (AI-4-EB) award, Grant BB/W013770/1. Mingzhu Shen acknowledges the funding from the Imperial College London President's PhD Scholarship. Furthermore, the authors would also like to thank Jing Zeng for her careful proofreading of this manuscript.
\end{ack}



\bibliography{mybibfile}


\clearpage
\appendix
\onecolumn

\section{Full Statistics of Evaluation Datasets} \label{appendix:all-graph-stats}

\begin{table*}[h!]
    \centering
    \caption{Statistics of all 26 datasets built for TopoAug evaluation.}
    \label{table:all-graph-stats}
    \resizebox{\textwidth}{!}{
    \begin{tabular}{llrrrrrrr}
        \toprule
        Name    & \makecell[l]{Hyperedge\\Construction\\Mechanism} & \makecell[r]{\#Nodes} & \makecell[r]{\#Edges} & \makecell[r]{\#Hyperedges} & \makecell[r]{Avg.\\Node\\Degree} & \makecell[r]{Avg.\\Hyperedge\\Degree} & \makecell[r]{\#Node\\Features} & \makecell[r]{\#Classes} \\
        \midrule
        MUSAE-GitHub & Graph Statistics & 37,700 & 578,006 & 223,672 & 30.66 & 4.591 & 4,005 or 128 & 4 \\ 
        MUSAE-Facebook & Graph Statistics & 22,470 & 342,004 & 236,663 & 30.44 & 9.905 & 4,714 or 128 & 4 \\
        MUSAE-TwitchDE & Graph Statistics & 9,498 & 306,276 & 297,315 & 64.49 & 7.661 & 3,170 or 128 & 2 \\ 
        MUSAE-TwitchEN & Graph Statistics & 7,126 & 70,648 & 13,248 & 19.83 & 3.666 & 3,170 or 128 & 2 \\  
        MUSAE-TwitchES & Graph Statistics & 4,648 & 118,764 & 77,135 & 51.10 & 5.826 & 3,170 or 128 & 2 \\ 
        MUSAE-TwitchFR & Graph Statistics & 6,549 & 225,332 & 172,653 & 68.81 & 5.920 & 3,170 or 128 & 2 \\  
        MUSAE-TwitchPT & Graph Statistics & 1,912 & 62,598 & 74,830 & 65.48 & 7.933 & 3,170 or 128 & 2 \\  
        MUSAE-TwitchRU & Graph Statistics & 4,385 & 74,608 & 25,673 & 34.03 & 4.813 & 3,170 or 128 & 2 \\  
        MUSAE-Wiki-Chameleon & Graph Statistics & 2,277 & 62,742 & 14,650 & 55.11 & 7.744 & 3,132 or 128 & Regression \\ 
        MUSAE-Wiki-Crocodile & Graph Statistics & 11,631 & 341,546 & 121,431 & 58.73 & 4.761 & 13,183 or 128 & Regression \\ 
        MUSAE-Wiki-Squirrel & Graph Statistics & 5,201 & 396,706 & 220,678 & 152.55 & 30.735 & 3,148 or 128 & Regression \\
        \midrule
        GRAND-ArteryAorta & Multi-Perspective & 5,848 & 5,823 & 11,368 & 1.991 & 1.277 &4,651&3\\ 
        GRAND-ArteryCoronary & Multi-Perspective & 5,755 & 5,722 & 11,222 & 1.989 & 1.273 & 4,651 & 3 \\ 
        GRAND-Breast & Multi-Perspective & 5,921 & 5,910 & 11,400 & 1.996 & 1.281 & 4,651 & 3 \\ 
        GRAND-Brain & Multi-Perspective & 6,196 & 6,245 & 11,878 & 2.016 & 1.296 & 4,651 & 3 \\ 
        GRAND-Lung & Multi-Perspective & 6,119 & 6,160 & 11,760 & 2.013 & 1.291 & 4,651 & 3 \\ 
        GRAND-Stomach & Multi-Perspective & 5,745 & 5,694 & 11,201 & 1.982 & 1.274 & 4,651 & 3 \\
        GRAND-Leukemia & Multi-Perspective & 4,651 & 6,362 & 7,812 & 2.736 & 1.324 & 4,651 & 3 \\ 
        GRAND-LungCancer & Multi-Perspective & 4,896 & 6,995 & 8,179 & 2.857 & 1.334 & 4,651 & 3 \\
        GRAND-StomachCancer & Multi-Perspective & 4,518 & 6,051 & 7,611 & 2.679 & 1.312 & 4,651 & 3 \\
        GRAND-KidneyCancer & Multi-Perspective & 4,319 & 5,599 & 7,369 & 2.593 & 1.297 & 4,651 & 3 \\ 
        Cora-CoAuthorship & Multi-Perspective & 2,708 & 5,429 & 1,072 & 4.010 & 4.277 & 1,433 & 7 \\
        Cora-CoCitation & Multi-Perspective & 2,708 & 5,429 & 1,579 & 4.010 & 3.030 & 1,433 & 7 \\
        Pubmed-CoCitation & Multi-Perspective & 19,717 & 44,338 & 7,963 & 4.497 & 4.349 & 500 & 3 \\
        \midrule
        Amazon-Computers & Multi-Modality & 10,226 & 55,324 & 10,226 & 10.82 & 3.000 & 1,000 & 10 \\ 
        Amazon-Photos & Multi-Modality & 6,777 & 45,306 & 6,777 & 13.37 & 4.800 & 1,000 & 10 \\ 
        \bottomrule
    \end{tabular}}
\end{table*}

\twocolumn

\section{Dataset and Task Specifications} \label{appendix:dataset-details}

\subsection{MUSAE Datasets}

\subsubsection{MUSAE-GitHub} This is a large social network of GitHub developers introduced in MUSAE~\cite{Rozemberczki2021MUSAE}. Nodes represent developers on GitHub, and edges are mutual follower relationships. TopoAug constructs the virtual hyperedges as mutually following developer groups that contain at least 3 developers (i.e., maximal cliques with sizes of at least 3). We enable the option to use either the 4,005-dimensional raw node features extracted based on the location, repositories starred, employer and e-mail address, or the 128-dimensional preprocessed node embeddings by MUSAE. The task is to predict whether a user is a web or a machine learning developer (could also be both or neither). 

\subsubsection{MUSAE-Facebook} This is a large Facebook page-page network introduced in MUSAE. Nodes represent verified pages on Facebook, and edges are mutual likes. TopoAug constructs the virtual hyperedges as mutually liked page groups that contain at least 3 pages (i.e., maximal cliques with sizes of at least 3). We enable the option to use either the 4,714-dimensional raw node features extracted from the site descriptions, or the 128-dimensional preprocessed node embeddings by MUSAE. The task is to predict the category a page belongs to: politicians, governmental organizations, television shows and companies.

\subsubsection{MUSAE-Twitch} These are six small Twitch user-user networks introduced in MUSAE. Nodes represent gamers on Twitch, and edges are mutual follower relationships between them. TopoAug constructs the virtual hyperedges as mutually following user groups that contain at least 3 gamers (i.e., maximal cliques with sizes of at least 3). We enable the option to use either the 3,170-dimensional raw node features extracted based on the games played and liked, location and streaming habits, or the 128-dimensional preprocessed node embeddings by MUSAE. The task is to predict whether a user streams mature content. 

\subsubsection{MUSAE-Wiki} These are three Wikipedia page-page networks dataset introduced in MUSAE. Nodes represent articles, and edges represent mutual hyperlinks between them. TopoAug constructs the virtual hyperedges as mutually linked page groups that contain at least 3 pages (i.e., maximal cliques with sizes of at least 3). We enable each dataset to have an option to use either the raw node features extracted based on informative nouns appeared in the text of the Wikipedia articles, or the 128-dimensional preprocessed node embeddings by MUSAE. The task is to predict the average monthly traffic of the web page.

\subsection{GRAND Datasets}

\subsubsection{GRAND-Tissues} We select and build six gene regulatory networks in different tissues (artery aorta, artery coronary, breast, brain, lung, and stomach) from GRAND~\cite{ben2022grand}, a public database for gene regulation. Nodes represent gene regulatory elements~\cite{maston2006transcriptional} with three distinct types: protein-encoding gene, lncRNA gene~\citep{long2017lncrnas}, and other regulatory elements. Edges are regulatory effects between genes. We train a CNN~\citep{eraslan2019deep} and use it to take the gene sequence as input and create a 4,651-dimensional embedding for each node. TopoAug constructs the virtual hyperedges by grouping nearby genomic elements on the chromosomes, i.e., the genomic elements within 200k base pair distance are grouped as virtual hyperedges. The task is a multi-class classification of gene regulatory elements. 

\subsubsection{GRAND-Diseases} We select and build four gene regulatory networks in different genetic diseases (leukemia, lung cancer, stomach cancer, and kidney cancer) from GRAND, a public database for gene regulation. Nodes represent gene regulatory elements with three distinct types: protein-encoding gene, lncRNA gene, and other regulatory elements. Edges are regulatory effects between genes. We train a CNN and use it to take the gene sequence as input and create a 4,651-dimensional embedding for each node. TopoAug constructs the virtual hyperedges by grouping nearby genomic elements on the chromosomes, i.e., the genomic elements within 200k base pair distance are grouped as virtual hyperedges. The task is a multi-class classification of gene regulatory elements. 

\subsection{Amazon Datasets}

The Amazon Computers and Photos datasets are two e-commerce graphs based on the Amazon Product Reviews dataset~\cite{mcauley2015image,he2016ups,ni2019justifying}. Nodes represent products, and an edge between two products is established if a user buys or writes reviews for both products. Node features are extracted based on the textual description of the products. TopoAug constructs the virtual hyperedges through the following process: first, it feeds the raw images of the products into a CLIP~\cite{radford2021learning} classifier, obtaining a 512-dimensional feature embedding for each image to assist the clustering; it then groups the products whose image embeddings’ pairwise distances are within a certain threshold into the same virtual hyperedges. The task is to predict the sub-category of a product in a multi-class classification setup.
\section{Results} \label{appendix:results}

We evaluate the following models on all 26 datasets we build:

\begin{itemize}[itemsep=\baselineskip]
    \item Simple-GNN baselines: GCN, GAT, and GraphSAGE;
    \item Hyper-GNN baselines: HyperConv, ED-HNN;
    \item Simple-GNNs with TopoAug, using simple-GNNs as auxiliary functions (these are used as baselines of ablation studies, as described in Section~\ref{section:ablation-simple-vs-hyper}):
    \begin{itemize}
        \item GCN+TopoAug\textsubscript{(GAT)}
        \item GCN+TopoAug\textsubscript{(GraphSAGE)}
        \item GAT+TopoAug\textsubscript{(GraphSAGE)}
    \end{itemize}
    \item A hyper-GNN with TopoAug, using another hyper-GNN as the auxiliary function:
    \begin{itemize}
        \item HyperConv+TopoAug\textsubscript{(ED-HNN)}
    \end{itemize}
    This is intended to assess whether the auxiliary features produced by TopoAug can be used to augment themselves, and improve the node prediction performance using only hyperedge information;
    \item Simple-GNNs with TopoAug, using hyper-GNNs as auxiliary functions (these are the standard usage of TopoAug):
    \begin{itemize}
        \item GCN+TopoAug\textsubscript{(HyperConv)}
        \item GCN+TopoAug\textsubscript{(ED-HNN)}
        \item GAT+TopoAug\textsubscript{(HyperConv)}
        \item GAT+TopoAug\textsubscript{(ED-HNN)}
        \item GraphSAGE+TopoAug\textsubscript{(GraphSAGE)}
        \item GraphSAGE+TopoAug\textsubscript{(ED-HNN)}
    \end{itemize}
\end{itemize}
Based on the time and compute resource constraints, each experiment is repeated two to five times with different random seeds. The results are summarised in \Cref{table:hyperaug-facebook-github,table:hyperaug-twitch,table:hyperaug-tissues,table:hyperaug-diseases,table:hyperaug-citation,table:hyperaug-amazon,table:hyperaug-wiki}, organised by the dataset subgroups. The results clearly demonstrate that simple-GNNs with TopoAug using hyper-GNNs as auxiliary functions consistently outperform the vanilla GNN baselines and other TopoAug combinations, underscoring the superior efficacy of incorporating information from higher-order node relations into graph augmentation methods under the TopoAug framework. Among all the TopoAug methods, GraphSAGE+TopoAug\textsubscript{(HyperConv)} generally stands out as the most effective combination.


\footnotetext[1]{\label{gpl}\url{https://www.gnu.org/licenses/gpl-3.0.html}}
\footnotetext[2]{\label{cc-by-sa}\url{https://creativecommons.org/licenses/by-sa/4.0/}}
\footnotetext[3]{\label{amazon-licence}\url{https://s3.amazonaws.com/amazon-reviews-pds/LICENSE.txt}}

\section{Licence} \label{appendix:licence}

The raw data for the MUSAE datasets are licenced under the the GNU General Public Licence, version 3 (GPLv3)\footnotemark[1]. The raw data for the GRAND datasets are licenced under the Creative Commons Attribution-ShareAlike 4.0 International Public Licence (CC BY-SA 4.0)\footnotemark[2]. The raw data for the Amazon datasets are licenced under the Amazon Service licence\footnotemark[3]. Having carefully observed the licence requirements of all data sources and code dependencies, we apply the same licenses to our constructed datasets as those of the raw data respectively.

\section{Ethics Statement}

All datasets constructed in this project are generated from public open-source datasets, and the original raw data downloaded from the data sources do not contain any personally identifiable information or other sensitive contents. The node prediction tasks for the datasets constructed in this project are designed to ensure that they do not, by any means, lead to discriminations against any social groups. Therefore, we are not aware of any social or ethical concern of TopoAug. Since TopoAug is a general graph augmentation method for representation learning on complex graphs, we also do not foresee any direct application of TopoAug to malicious purposes. However, the community should be aware of any potential negative social and ethical impacts that may arise from their chosen datasets or tasks outside of those provided in this project.

\begin{table*}[htbp]
    \caption{Accuracy (\%) of the TopoAug variants and vanilla GNN baselines on the MUSAE-Facebook and GitHub datasets.}
    \label{table:hyperaug-facebook-github}
    \centering
    \begin{tabular}{lll}
        \toprule
        Method  & Facebook & GitHub \\
        \midrule
        RandomGuess & \textit{25.0} & \textit{25.0} \\
        \midrule
        GCN & 88.6 $\pm$ 0.1 & 87.2 $\pm$ 0.0 \\
        GAT & 87.6 $\pm$ 0.1 & 86.4 $\pm$ 0.1 \\
        GraphSAGE  & 90.2 $\pm$ 0.2 & 87.1 $\pm$ 0.2 \\
        HyperConv  & 79.2 $\pm$ 0.1 & 80.8 $\pm$ 0.1 \\
        ED-HNN & 86.1 $\pm$ 0.4 & 86.2 $\pm$ 0.1 \\
        \midrule
        GCN+TopoAug\textsubscript{(GAT)} & 91.0 $\pm$ 0.1 & 86.7 $\pm$ 0.1 \\
        GCN+TopoAug\textsubscript{(GraphSAGE)} & 93.8 $\pm$ 0.3 & 87.0 $\pm$ 0.4 \\
        GAT+TopoAug\textsubscript{(GraphSAGE)} & 93.7 $\pm$ 0.4 & 86.8 $\pm$ 0.6 \\
        HyperConv+TopoAug\textsubscript{(ED-HNN)} & \textbf{94.2 $\pm$ 0.1} & 86.9 $\pm$ 0.3 \\
        \midrule
        GCN+TopoAug\textsubscript{(HyperConv)} & 89.8 $\pm$ 0.0 & 87.2 $\pm$ 0.0 \\
        GCN+TopoAug\textsubscript{(ED-HNN)} & 92.6 $\pm$ 0.7 & \textbf{87.4 $\pm$ 0.3} \\
        GAT+TopoAug\textsubscript{(HyperConv)} & 90.5 $\pm$ 0.0 & 86.0 $\pm$ 0.2 \\
        GAT+TopoAug\textsubscript{(ED-HNN)} & 93.4 $\pm$ 0.1 & 87.3 $\pm$ 0.1 \\
        GraphSAGE+TopoAug\textsubscript{(HyperConv)} & 93.6 $\pm$ 0.0 & 87.3 $\pm$ 0.2 \\
        GraphSAGE+TopoAug\textsubscript{(ED-HNN)} & 93.2 $\pm$ 0.7 & \textbf{87.4 $\pm$ 0.1} \\
        \bottomrule
    \end{tabular}
\end{table*}

\begin{table*}[htbp]
    \caption{Accuracy (\%) of the TopoAug variants and vanilla GNN baselines on the MUSAE-Twitch datasets.}
    \label{table:hyperaug-twitch}
    \centering
    \begin{tabular}{lllllll}
        \toprule
        Method & TwitchES & TwitchFR & TwitchDE & TwitchEN & TwitchPT & TwitchRU \\
        \midrule
        RandomGuess & \textit{50.0} & \textit{50.0} & \textit{50.0} & \textit{50.0} & \textit{50.0} & \textit{50.0} \\
        \midrule
        GCN & 72.1 $\pm$ 0.4 & 62.4 $\pm$ 0.1 & 65.5 $\pm$ 0.2 & 62.0 $\pm$ 0.3 & 68.9 $\pm$ 0.6 & 74.5 $\pm$ 0.0 \\
        GAT & 69.4 $\pm$ 0.2 & 62.3 $\pm$ 0.0 & 64.5 $\pm$ 0.4 & 59.4 $\pm$ 0.6 & 66.4 $\pm$ 0.7 & 74.3 $\pm$ 0.2 \\
        GraphSAGE & 69.0 $\pm$ 0.2 & 61.6 $\pm$ 0.3 & 65.7 $\pm$ 0.1 & 60.5 $\pm$ 0.0 & 67.2 $\pm$ 1.3 & 74.5 $\pm$ 0.1 \\
        HyperConv & 71.5 $\pm$ 0.1 & 62.4 $\pm$ 0.2 & 65.4 $\pm$ 0.2 & 58.7 $\pm$ 0.7 & 70.1 $\pm$ 0.5 & 74.1 $\pm$ 0.1 \\
        ED-HNN & 72.2 $\pm$ 1.3 & 62.9 $\pm$ 1.1 & 68.1 $\pm$ 0.6 & 60.3 $\pm$ 1.1 & 69.5 $\pm$ 1.5 & 75.1 $\pm$ 1.5 \\
        \midrule
        GCN+TopoAug\textsubscript{(GAT)} & 72.7 $\pm$ 0.1 & 62.3 $\pm$ 0.1 & 66.2 $\pm$ 0.1 & 61.2 $\pm$ 0.2 & 68.6 $\pm$ 0.8 & 74.5 $\pm$ 0.1 \\
        GCN+TopoAug\textsubscript{(GraphSAGE)} & 71.8 $\pm$ 2.5 & 61.2 $\pm$ 0.0 & 68.2 $\pm$ 1.6 & 59.2 $\pm$ 2.1 & 70.0 $\pm$ 1.9 & 74.8 $\pm$ 0.0 \\
        GAT+TopoAug\textsubscript{(GraphSAGE)} & 70.8 $\pm$ 0.7 & 61.4 $\pm$ 0.0 & 67.7 $\pm$ 0.5 & 59.3 $\pm$ 0.2 & 68.0 $\pm$ 2.1 & 74.9 $\pm$ 0.4 \\
        HyperConv+TopoAug\textsubscript{(ED-HNN)} & 71.8 $\pm$ 1.0 & 61.2 $\pm$ 0.3 & 68.3 $\pm$ 0.3 & 59.3 $\pm$ 0.7 & 67.9 $\pm$ 0.2 & 75.1 $\pm$ 0.3 \\
        \midrule
        GCN+TopoAug\textsubscript{(HyperConv)} & \textbf{72.9 $\pm$ 0.1} & 62.6 $\pm$ 0.0 & 65.7 $\pm$ 0.1 & 60.7 $\pm$ 0.1 & 69.6 $\pm$ 0.4 & 74.4 $\pm$ 0.0 \\
        GCN+TopoAug\textsubscript{(ED-HNN)} & 72.8 $\pm$ 1.4 & \textbf{63.4 $\pm$ 2.1} & 67.9 $\pm$ 0.9 & \textbf{62.2 $\pm$ 1.0 } & \textbf{70.6 $\pm$ 3.7} & 75.2 $\pm$ 0.4 \\
        GAT+TopoAug\textsubscript{(HyperConv)} & 71.4 $\pm$ 0.1 & 62.2 $\pm$ 0.0 & 65.4 $\pm$ 0.1 & 60.8 $\pm$ 0.3 & 67.2 $\pm$ 0.3 & 74.4 $\pm$ 0.0 \\
        GAT+TopoAug\textsubscript{(ED-HNN)} & 72.1 $\pm$ 2.3 & 63.2 $\pm$ 3.1 & \textbf{68.5 $\pm$ 0.2} & 61.0 $\pm$ 0.6 & 68.7 $\pm$ 2.8 & \textbf{75.3 $\pm$ 0.2} \\
        GraphSAGE+TopoAug\textsubscript{(HyperConv)} & 71.8 $\pm$ 1.0 & 63.3 $\pm$ 2.2 & 68.3 $\pm$ 0.3 & 61.1 $\pm$ 0.2 & 69.9 $\pm$ 2.4 & 75.1 $\pm$ 0.3 \\
        GraphSAGE+TopoAug\textsubscript{(ED-HNN)} & 70.5 $\pm$ 0.8 & 63.3 $\pm$ 2.4 & 68.1 $\pm$ 0.8 & 60.7 $\pm$ 0.0 & 68.9 $\pm$ 0.2 & 75.2 $\pm$ 0.3 \\
        \bottomrule
    \end{tabular}
\end{table*}


\begin{table*}[htbp]
    \caption{Accuracy (\%) of the TopoAug variants and vanilla GNN baselines on the GRAND-Tissues datasets.}
    \label{table:hyperaug-tissues}
    \centering
    \begin{tabular}{lllllll}
        \toprule
        Method & \makecell[l]{Artery\\Aorta} & \makecell[l]{Artery\\Coronary} & Breast & Brain & Lung & Stomach \\
        \midrule
        RandomGuess & \textit{33.3} & \textit{33.3} & \textit{33.3} & \textit{33.3} & \textit{33.3} & \textit{33.3} \\
        \midrule
        GCN & 62.7 $\pm$ 0.7 & 66.2 $\pm$ 0.1 & 63.9 $\pm$ 1.0 & 62.5 $\pm$ 0.0 & 65.0 $\pm$ 0.0 & 64.3 $\pm$ 0.0 \\
        GAT & 62.8 $\pm$ 0.4 & 66.3 $\pm$ 0.0 & 64.3 $\pm$ 0.1 & 62.5 $\pm$ 0.1 & 64.8 $\pm$ 0.4 & 64.3 $\pm$ 0.0 \\
        GraphSAGE & 62.8 $\pm$ 0.2 & 66.3 $\pm$ 0.1 & 64.4 $\pm$ 0.0 & 61.8 $\pm$ 0.2 & 64.6 $\pm$ 0.5 & 63.0 $\pm$ 1.0 \\
        HyperConv & 62.6 $\pm$ 0.8 & 66.2 $\pm$ 0.0 & 64.5 $\pm$ 0.1 & 62.5 $\pm$ 0.0 & 65.0 $\pm$ 0.0 & 64.3 $\pm$ 0.0 \\
        ED-HNN & 66.7 $\pm$ 1.3 & 66.7 $\pm$ 1.5 & 66.1 $\pm$ 1.4 & 66.3 $\pm$ 1.3 & 67.0 $\pm$ 1.0 & 66.1 $\pm$ 0.6 \\
        \midrule
        GCN+TopoAug\textsubscript{(GAT)} & 62.7 $\pm$ 0.0 & 64.1 $\pm$ 0.0 & 62.6 $\pm$ 0.1 & 62.5 $\pm$ 0.3 & 62.7 $\pm$ 0.1 & 62.6 $\pm$ 0.1 \\
        GCN+TopoAug\textsubscript{(GraphSAGE)} & 64.6 $\pm$ 0.3 & 65.9 $\pm$ 2.2 & 65.3 $\pm$ 0.3 & 65.3 $\pm$ 0.9 & 64.0 $\pm$ 0.1 & 63.8 $\pm$ 1.6 \\
        GAT+TopoAug\textsubscript{(GraphSAGE)} & 66.4 $\pm$ 0.1 & 66.3 $\pm$ 1.8 & 65.4 $\pm$ 0.2 & 65.1 $\pm$ 0.8 & 65.0 $\pm$ 0.1 & 65.0 $\pm$ 2.2 \\
        HyperConv+TopoAug\textsubscript{(ED-HNN)} & 65.9 $\pm$ 0.0 & 66.2 $\pm$ 1.1 & 66.7 $\pm$ 1.0 & 66.1 $\pm$ 1.6 & 67.0 $\pm$ 0.7 & 66.8 $\pm$ 2.7 \\
        \midrule
        GCN+TopoAug\textsubscript{(HyperConv)} & 64.7 $\pm$ 0.3 & 66.0 $\pm$ 0.3 & 65.2 $\pm$ 0.6 & 63.7 $\pm$ 0.2 & 65.4 $\pm$ 0.4 & 65.4 $\pm$ 0.1 \\
        GCN+TopoAug\textsubscript{(ED-HNN)} & 64.3 $\pm$ 0.2 & 66.0 $\pm$ 2.0 & 66.1 $\pm$ 1.0 & \textbf{66.7 $\pm$ 2.6} & 67.9 $\pm$ 1.3 & 66.9 $\pm$ 0.0 \\
        GAT+TopoAug\textsubscript{(HyperConv)} & 64.9 $\pm$ 0.1 & 66.4 $\pm$ 0.1 & 65.7 $\pm$ 0.1 & 64.5 $\pm$ 0.1 & 66.2 $\pm$ 0.2 & 65.6 $\pm$ 0.2 \\
        GAT+TopoAug\textsubscript{(ED-HNN)} & 65.3 $\pm$ 2.0 & 66.3 $\pm$ 1.1 & 65.6 $\pm$ 1.7 & 65.5 $\pm$ 2.4 & \textbf{68.3 $\pm$ 0.7} & 66.3 $\pm$ 0.8 \\
        GraphSAGE+TopoAug\textsubscript{(HyperConv)} & \textbf{66.9 $\pm$ 0.9} & \textbf{67.9 $\pm$ 0.0} & \textbf{67.5 $\pm$ 0.6} & 66.6 $\pm$ 1.6 & 67.4 $\pm$ 0.7 & \textbf{68.1 $\pm$ 0.3} \\
        GraphSAGE+TopoAug\textsubscript{(ED-HNN)} & 66.2 $\pm$ 0.5 & 67.1 $\pm$ 2.5 & 64.7 $\pm$ 0.8 & 66.0 $\pm$ 0.2 & 67.9 $\pm$ 0.5 & 66.9 $\pm$ 0.5 \\
        \bottomrule
    \end{tabular}
\end{table*}

\begin{table*}[htbp]
    \caption{Accuracy (\%) of the TopoAug variants and vanilla GNN baselines on the GRAND-Diseases datasets.}
    \label{table:hyperaug-diseases}
    \centering
    \begin{tabular}{lllll}
        \toprule
        Method & Leukemia & \makecell[l]{Lung\\Cancer} & \makecell[l]{Stomach\\Cancer} & \makecell[l]{Kidney\\Cancer} \\
        \midrule
        RandomGuess & \textit{33.3} & \textit{33.3} & \textit{33.3} & \textit{33.3} \\
        \midrule
        GCN & 58.2 $\pm$ 0.1 & 59.6 $\pm$ 0.1 & 60.2 $\pm$ 0.7 & 58.1 $\pm$ 0.2 \\
        GAT & 58.7 $\pm$ 0.5 & 59.6 $\pm$ 0.0 & 60.0 $\pm$ 0.7 & 58.1 $\pm$ 0.3 \\
        GraphSAGE & 60.4 $\pm$ 1.6 & 61.5 $\pm$ 1.5 & 60.2 $\pm$ 1.6 & 59.6 $\pm$ 1.4 \\
        HyperConv & 58.6 $\pm$ 0.3 & 59.3 $\pm$ 0.3 & 59.6 $\pm$ 0.4 & 57.7 $\pm$ 0.6 \\
        ED-HNN & 60.3 $\pm$ 1.3 & 60.2 $\pm$ 1.4 & 61.0 $\pm$ 1.1 & 60.3 $\pm$ 1.1 \\
        \midrule
        GCN+TopoAug\textsubscript{(GAT)} & 59.0 $\pm$ 0.2 & 58.3 $\pm$ 0.2 & 58.1 $\pm$ 0.4 & 58.4 $\pm$ 0.2 \\
        GCN+TopoAug\textsubscript{(GraphSAGE)} & 60.1 $\pm$ 0.2 & 64.1 $\pm$ 0.2 & 62.1 $\pm$ 0.2 & 62.1 $\pm$ 0.2 \\
        GAT+TopoAug\textsubscript{(GraphSAGE)} & 61.8 $\pm$ 0.5 & 63.2 $\pm$ 0.7 & 62.4 $\pm$ 1.9 & 61.2 $\pm$ 1.1 \\
        HyperConv+TopoAug\textsubscript{(ED-HNN)} & 58.9 $\pm$ 1.4 & 61.5 $\pm$ 1.0 & 62.3 $\pm$ 2.1 & 61.1 $\pm$ 0.3 \\
        \midrule
        GCN+TopoAug\textsubscript{(HyperConv)} & 60.4 $\pm$ 0.4 & 61.4 $\pm$ 0.6 & 60.5 $\pm$ 0.3 & 60.9 $\pm$ 0.4 \\
        GCN+TopoAug\textsubscript{(ED-HNN)} & 61.5 $\pm$ 1.0 & 63.7 $\pm$ 0.9 & 63.4 $\pm$ 2.0 & 62.4 $\pm$ 1.3 \\
        GAT+TopoAug\textsubscript{(HyperConv)} & 62.5 $\pm$ 1.2 & 61.8 $\pm$ 0.3 & 64.2 $\pm$ 1.6 & 64.0 $\pm$ 0.3 \\
        GAT+TopoAug\textsubscript{(ED-HNN)} & 62.3 $\pm$ 1.2 & 61.5 $\pm$ 1.2 & 64.5 $\pm$ 1.9 & 63.8 $\pm$ 1.8 \\
        GraphSAGE+TopoAug\textsubscript{(HyperConv)} & \textbf{62.9 $\pm$ 0.5} & \textbf{66.4 $\pm$ 0.2} & \textbf{66.3 $\pm$ 1.5} & \textbf{64.8 $\pm$ 0.4} \\
        GraphSAGE+TopoAug\textsubscript{(ED-HNN)} & 62.6 $\pm$ 0.3 & 64.5 $\pm$ 0.1 & 64.7 $\pm$ 0.7 & 63.9 $\pm$ 0.2 \\
        \bottomrule
    \end{tabular}
\end{table*}

\begin{table*}[htbp]
    \caption{Accuracy (\%) of the TopoAug variants and vanilla GNN baselines on the citation datasets. *The results for GCN and GAT are directly taken from the GAT paper~\cite{velivckovic2017graph}. **The results for HyperConv and ED-HNN are directly taken from the ED-HNN paper~\cite{Wang2023EDHNN}.}
    \label{table:hyperaug-citation}
    \centering
    \begin{tabular}{llll}
        \toprule
        Method & Cora-CoCite & Cora-CoAuth & Pubmed-CoCite \\
        \midrule
        RandomGuess & \textit{14.3} & \textit{14.3} & \textit{33.3} \\
        \midrule
        GCN* & 81.4 $\pm$ 0.5 & 81.4 $\pm$ 0.5 & 79.0 $\pm$ 0.3 \\
        GAT* & 83.0 $\pm$ 0.7 & 83.0 $\pm$ 0.7 & 79.0 $\pm$ 0.3 \\
        GraphSAGE & 83.2 $\pm$ 0.1 & 83.2 $\pm$ 0.1 & 86.5 $\pm$ 0.8 \\
        HyperConv** & 79.1 $\pm$ 1.0 & 82.6 $\pm$ 1.0 & 86.4 $\pm$ 0.4 \\
        ED-HNN** & 80.3 $\pm$ 1.4 & 84.0 $\pm$ 1.6 & \textbf{89.0 $\pm$ 0.5} \\
        \midrule
        GCN+TopoAug\textsubscript{(GAT)} & 86.9 $\pm$ 0.6 & 86.9 $\pm$ 0.6 & 87.0 $\pm$ 1.0 \\
        GCN+TopoAug\textsubscript{(GraphSAGE)} & \textbf{87.4 $\pm$ 1.4} & 87.4 $\pm$ 1.4 & 87.4 $\pm$ 0.6 \\
        GAT+TopoAug\textsubscript{(GraphSAGE)} & 85.8 $\pm$ 0.0 & 85.8 $\pm$ 0.0 & 87.3 $\pm$ 1.0 \\
        HyperConv+TopoAug\textsubscript{(ED-HNN)} & 83.0 $\pm$ 1.0 & 85.2 $\pm$ 1.4 & 87.4 $\pm$ 0.1 \\
        \midrule
        GCN+TopoAug\textsubscript{(HyperConv)} & 86.9 $\pm$ 1.7 & 87.5 $\pm$ 0.9 & 87.3 $\pm$ 0.4 \\
        GCN+TopoAug\textsubscript{(ED-HNN)} & 86.6 $\pm$ 1.4 & 87.0 $\pm$ 0.8 & \textbf{89.0 $\pm$ 0.2} \\
        GAT+TopoAug\textsubscript{(HyperConv)} & 86.7 $\pm$ 0.4 & 86.0 $\pm$ 2.2 & 86.6 $\pm$ 0.7 \\
        GAT+TopoAug\textsubscript{(ED-HNN)} & 87.0 $\pm$ 1.3 & 86.7 $\pm$ 0.0 & \textbf{89.0 $\pm$ 0.6} \\
        GraphSAGE+TopoAug\textsubscript{(HyperConv)} & 87.2 $\pm$ 1.8 & 86.3 $\pm$ 0.8 & 87.3 $\pm$ 0.4 \\
        GraphSAGE+TopoAug\textsubscript{(ED-HNN)} & \textbf{87.4 $\pm$ 1.0} & \textbf{87.7 $\pm$ 0.8} & 88.8 $\pm$ 1.1 \\
        \bottomrule
    \end{tabular}
\end{table*}

\begin{table*}[htbp]
    \caption{Accuracy (\%) of the TopoAug variants and vanilla GNN baselines on the Amazon datasets.}
    \label{table:hyperaug-amazon}
    \centering
    \begin{tabular}{lll}
        \toprule
        Method & Computers & Photos \\
        \midrule
        RandomGuess & \textit{10.0} & \textit{10.0} \\
        \midrule
        GCN & 75.6 $\pm$ 4.1 & 29.5 $\pm$ 1.7 \\
        GAT & 74.2 $\pm$ 4.3 & 43.4 $\pm$ 7.4 \\
        GraphSAGE & 75.0 $\pm$ 1.9 & 36.6 $\pm$ 6.1 \\
        HyperConv & 84.2 $\pm$ 2.0 & 33.7 $\pm$ 5.9 \\
        ED-HNN & 97.3 $\pm$ 0.2 & 78.6 $\pm$ 1.2 \\
        \midrule
        GCN+TopoAug\textsubscript{(GAT)} & 91.3 $\pm$ 0.1 & 71.1 $\pm$ 0.5 \\
        GCN+TopoAug\textsubscript{(GraphSAGE)} & 93.0 $\pm$ 0.0 & 78.6 $\pm$ 0.6 \\
        GAT+TopoAug\textsubscript{(GraphSAGE)} & 96.8 $\pm$ 0.6 & 78.0 $\pm$ 0.1 \\
        HyperConv+TopoAug\textsubscript{(ED-HNN)} & 97.8 $\pm$ 0.1 & 80.7 $\pm$ 0.6 \\
        \midrule
        GCN+TopoAug\textsubscript{(HyperConv)} & 96.8 $\pm$ 0.5 & 69.8 $\pm$ 0.5 \\
        GCN+TopoAug\textsubscript{(ED-HNN)} & 98.1 $\pm$ 0.7 & 80.5 $\pm$ 0.6 \\
        GAT+TopoAug\textsubscript{(HyperConv)} & 96.8 $\pm$ 0.6 & 71.5 $\pm$ 0.3 \\
        GAT+TopoAug\textsubscript{(ED-HNN)} & \textbf{98.2 $\pm$ 0.1} & \textbf{80.9 $\pm$ 0.3} \\
        GraphSAGE+TopoAug\textsubscript{(HyperConv)} & \textbf{98.2 $\pm$ 0.7} & 79.3 $\pm$ 0.8 \\
        GraphSAGE+TopoAug\textsubscript{(ED-HNN)} & 98.1 $\pm$ 0.1 & \textbf{80.9 $\pm$ 0.7} \\
        \bottomrule
    \end{tabular}
\end{table*}


\end{document}